\documentclass[12pt]{article}

\usepackage{
    amsfonts,
    amsmath,
    amssymb,
    array,
    booktabs,
    caption,
    caption,
    color,
    calc,
    comment,
    eurosym,
    fancyhdr,
    float,
    footmisc,
    geometry,
    graphicx,
    hyperref,
    natbib,
    pdflscape,
    sectsty,
    setspace,
    subfig,
    ulem,
}

\newcolumntype{L}[1]{>{\raggedright\let\newline\\arraybackslash\hspace{0pt}}m{#1}}
\newcolumntype{C}[1]{>{\centering\let\newline\\arraybackslash\hspace{0pt}}m{#1}}
\newcolumntype{R}[1]{>{\raggedleft\let\newline\\arraybackslash\hspace{0pt}}m{#1}}

\begin{document}

\begin{titlepage}
\title{Can We Trust Race Prediction?}
\author{Cangyuan Li\thanks{{I provide a Python package at \url{https://github.com/CangyuanLi/pyethnicity}. Active model development can be found at \url{https://github.com/CangyuanLi/race_ml}. The author can be contacted at cangyuan.c.li@gmail.com.}}}
\date{\today}
\maketitle
\begin{abstract}
    \noindent 

    In the absence of sensitive race and ethnicity data, researchers, regulators, and firms alike turn to proxies. In this paper, I train a Bidirectional Long Short-Term Memory (BiLSTM) model on a novel dataset of voter registration data from all 50 US states and create an ensemble that achieves up to 36.8\% higher out of sample (OOS) F1 scores than the best performing machine learning models in the literature. Additionally, I construct the most comprehensive database of first and surname distributions in the US in order to improve the coverage and accuracy of Bayesian Improved Surname Geocoding (BISG) and Bayesian Improved Firstname Surname Geocoding (BIFSG). Finally, I provide the first high-quality benchmark dataset in order to fairly compare existing models and aid future model developers.

    \vspace{0.2in}

    \bigskip
\end{abstract}
\setcounter{page}{0}
\thispagestyle{empty}
\end{titlepage}
\pagebreak \newpage

\doublespacing

\section{Introduction} \label{sec:introduction}

Race prediction has wide applications across a diverse range of fields, from lending to criminal justice to healthcare.
As race and ethnicity are sensitive pieces of information, many datasets, public and private, do not have access to
``true'' race, and must therefore rely on proxies. For example, the Consumer Finance Protection Bureau (CFPB) uses Bayesian Improved Surname Geocoding (BISG), which relies on name and zip code, in their fair lending analysis \cite{cfpb2014}. Race prediction is essential to research involving racial outcomes, such as in \citep{brown2016,frame2022,clifford2023}. However, there are necessary trade-offs between coverage, accuracy, and ease of use. For example, \cite{sood2018} provide surname-only models in addition to their full name models despite the Census Bureau's last name database being the Bayes optimal solution in order to improve coverage (for instance, if a name is not in the database). \cite{greenwald2023} exploit the fact that image-based models such as Facenet512 achieve accuracies above 99\% (even better than humans at \( \sim \)98\%) to create an image-based measure of race that performs much better than BISG. However, datasets with headshots are rare, and it may not be feasible for some applications to collect such images. Even when images are unavailable, it is likely that more granular features such as income, address, and political affiliation could improve race prediction. Yet, there are many contexts in which such data is simply unavailable. In this paper, in order to maximize generalizability, I consider the most common context in which race prediction is used--when only some combination of name and zip code is available--and provide an ensemble model that outperforms all open-source solutions I was able to test.

\begin{figure}[H]
    \caption{ROC Curve Comparison}
    \label{f:roc_compare}
    \centering
    \includegraphics[scale=.95]{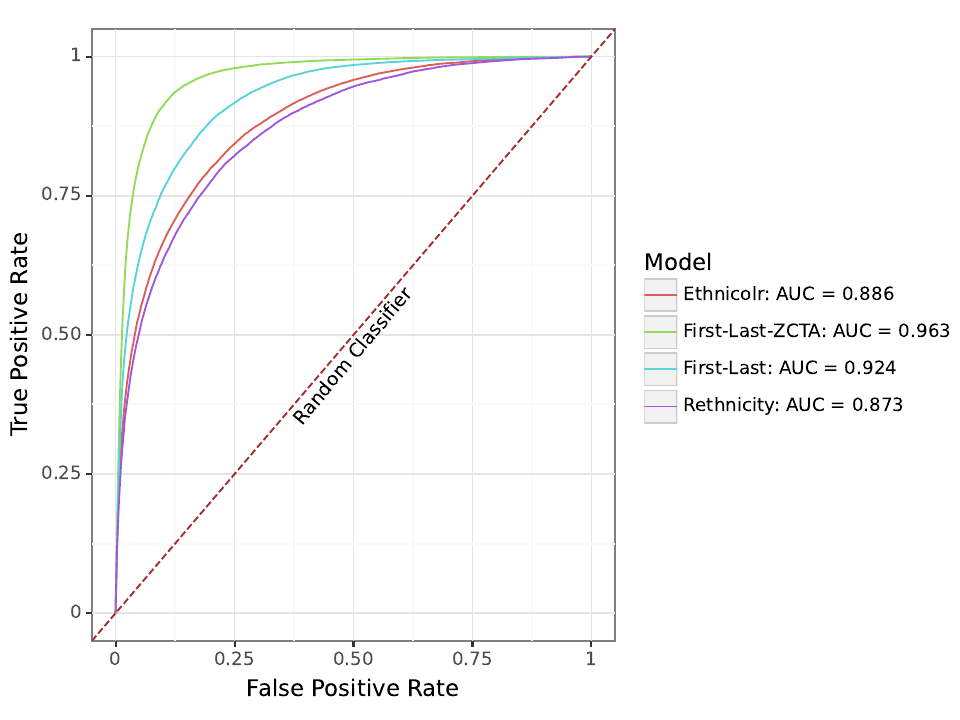}
\end{figure}

I begin by training a Bidirectional Long Short-Term Memory (BiLSTM) model on voter registration data to predict race from first and last name alone. While voter registration data has been used before (\citep{sood2018,fang2022,imai2016}, to name just a few), the data has thus far been limited to some combination of six southern US states (Alabama, Florida, Georgia, Louisiana, North Carolina, South Carolina, Tennessee, and Texas), with many papers using only Florida data. I source 2023 voter registration data from all 50 US states, representing the largest and most geographically diverse corpus of name and race in the literature. The First-Last model obtains out of sample (OOS) F1 scores up to 16.3\% higher than the next-best performing machine learning model. When combined with zip code tabulation area (ZCTA) features using naive Bayes (First-Last-ZCTA), the number jumps to 35.3\%. Figure \ref{f:roc_compare} compares the receiver operating characteristic curves for Black (consistently the most difficult class to predict) of First-Last and First-Last-ZCTA with available open-source models, namely ethnicolr (\cite{sood2018}) and rethnicity (\cite{fang2022}).

While BiLSTM shows good performance, it is also powerful to be able to empirically calculate the racial distribution of a name. Indeed, if one had perfect information, the optimal prediction for a given name and geography is simply the race that has the greatest percent of people in that geography with that name. Data limitations make such an approach impossible, but BISG (\cite{elliot2009}) and Bayesian Improved Firstname Surname Geocoding (BIFSG; \cite{voicu2018}) provide good approximations. In particular, BISG has become the de facto standard race prediction algorithm for its coverage, interpretability, and solid performance. BIFSG seeks to improve on BISG, especially for Blacks (who often have similar surnames as Whites but distinctive first names (\cite{bertrand2004})), by incorporating first name data. However, the existing first name list provided by \cite{tzioumis2018} comes from a comparatively small sample of 2,663,364 observations representing 91,526 unique first names sourced from Home Mortgage Disclosure Act (HMDA) data. As such, it can be prone to coverage and performance issues. On a test sample of 200,000 Paycheck Protection Program (PPP) borrowers, for Black, BIFSG has 43.4\% worse coverage and a 3.8\% lower F1 score compared to BISG. I significantly expand on this selection and provide 51,294,450 observations representing 1,187,576 distinct first names, the most comprehensive first name database to date. Compared to the old data, the updated data results in up to 46.6\% better coverage and 12.5\% higher F1 scores.

In order to fairly examine and compare the performance of the aforementioned algorithms, a benchmark dataset is crucial. Standard metrics such as precision, recall, and F1 are highly sample-dependent. One paper may validate against a sample of Florida voter registration data (\cite{sood2018}), another on a national sample of proprietary loan data (\cite{kotova2021}), each with varying racial and geographic distributions. Additionally, since there are only a few viable training datasets, one paper's testing dataset may contain examples found in another's training dataset, making direct comparisons even more difficult. To address this issue, I use publicly available PPP data containing self-reported race (the gold standard for ground truth race) to create a large database of 1,066,605 unique name / zip code tabulation area (ZCTA) pairs. To the best of my knowledge, no existing model trains on this data. Additionally, it is geographically diverse, containing loans from 57 US states / territories and 27,702 out of 33,121 (84.6\%) ZCTAs. I use the PPP data to conduct a replicable, ``fair'' comparison of various race prediction algorithms.

\subsection{Definitions}

Throughout the paper, I use several metrics to assess the performance of different models. For each class, I calculate the number of True Positives (e.g. the predicted and self-reported races are both Asian), True Negatives (e.g. the predicted race is not Asian and the self-reported race is not Asian), False Positives (e.g. the predicted race is Asian but the self-reported race is not Asian), and False Negatives (e.g. the predicted race is not Asian but the self-reported race is Asian). Since each model returns the probability a person is of a certain race, I determine the predicted race by taking the maximum of the probabilities, which I call ``Max.'' The following table provides a summary of the metrics used.

\begin{table}[H]
    \centering
    \begin{tabular}{@{}ccc@{}}
    \toprule
    Metric   & Formula   & Interpretation                                    \\ \midrule
    Accuracy & \( \frac{TP + TN}{TP + TN + FP + FN} \) & Ratio of correct predictions to total predictions \\
    Precision & \( \frac{TP}{TP + FP}  \) & Percent of correct positive predictions \\
    Recall & \( \frac{TP}{TP + FN} \) & Percent of actual positives identified \\
    F1 Score & \( \frac{2 \times Precision \times Recall}{Precision + Recall} \) & Harmonic mean of precision and recall \\
    Support & & The number of samples that have a valid prediction \\
    Coverage & & The percentage of samples that have a valid prediction \\
    \bottomrule
    \end{tabular}
\end{table}

\section{Literature Review} \label{sec:literature}

This paper contributes to the race prediction literature, both in terms of model development and evaluation. \cite{sood2018} and \cite{fang2022} both use Florida voter registration data to train an LSTM to predict race from first and last name. \cite{voicu2018} improves upon the BISG algorithm introduced by \cite{elliot2009} by adding first name data. This paper's key contributions are to significantly improve the coverage and accuracy of BIFSG, provide a state-of-the-art machine learning model to fill in the gaps left by BIFSG, and make available a clean, nationally representative dataset for model developers to benchmark against.

\section{Data} \label{sec:data}

\subsection{L2}

I use two primary datasets in the paper. The first is 2023 voter registration data sourced from L2 Inc., one of the leading providers of voter data in the US. The data spans 58 US states / territories, representing 32,034 out of 33,121 (96.7\%) unique ZCTAs. I filter to the four major race / ethnicity categories: Non-Hispanic Asian, Non-Hispanic Black, Hispanic, and Non-Hispanic White. When self-reported race is not directly available, I use the ethnicity field, and follow the guidelines used by the Office of Management and Budget and the Census Bureau. For example, ``White'' includes people who report their ethnicity as German, Irish, English, etc., whereas ``Asian'' includes people who originate from countries such as China, India, and Japan (\cite{racedefs}). Table \ref{t:l2_racial_distribution} provides a breakdown.

\begin{table}[h]
    \caption{L2 Racial Distribution}
    \label{t:l2_racial_distribution}
    \centering
    \begin{tabular}{llll}
\toprule
Race & Total & Self-Reported & From Ethnicity \\
\midrule
Asian & 7,722,316 & 575,210 & 7,147,106 \\
Black & 6,681,001 & 6,681,001 & 0 \\
Hispanic & 23,758,058 & 493,213 & 23,264,845 \\
White & 120,272,305 & 17,774,105 & 102,498,200 \\
\bottomrule
\end{tabular}

    \begin{minipage}{\textwidth} 
        \medskip
		\footnotesize{
            \textbf{Note:} L2 combines the ``African'' ethnicity and ``African-American Self-Reported'' into one field, which I call ``Black.''
        }
	\end{minipage}
\end{table}

The L2 data is used to train the BiLSTM and build the expanded first and surname distributions.

\subsection{PPP}

To address the need for a benchmark dataset in the literature, I use PPP data to build a nationally representative database of name, geography (ZCTA), and self-reported race. I begin with a dataset of 11,460,475 loans spanning April 3, 2020 to May 31, 2021 and subset to the 1,211,770 name / zip pairs that both self-report race and are person names. Non-person names are removed using a custom list of over 1,000 filter words, such as ``llc'', ``installation'', and so on. The final dataset represents 1,066,605 unique name / ZCTA pairs, 27,702 out of 33,121 (84.6\%) ZCTAs, and 57 states / territories.

Conducting a fair comparison is harder than it may first appear. The ultimate goal of these models is to perform in a ``real-world'' setting. However, there are many equally valid ``real-world'' settings, and model performance can oscillate wildly between different distributions. For example, there is often significant overlap between Black and White names. A name such as ``Dorothy Brown'' encodes very little information about whether that person is Black or White, even to a human. Therefore, if Model X overpredicts Black, it would perform well on a sample where Black is the majority class, but would exhibit high false positive rates on a sample that is majority White. Concretely, the PPP sample is a real-word setting, but it is somewhat selected. Instead of a random draw from the entire US population, it is a sample of small business owners who received a PPP loan. Additionally, since the Small Business Administration prioritized lending to minority-owned businesses in the final months of the program, the dataset is majority Black, which is not true of the general US. To address this issue, I assume that the target distribution roughly reflects the US population, and draw a nationally representative sample of 200,000 observations\footnote{I use the July 1, 2022, population estimates from \url{https://www.census.gov/quickfacts/fact/table/US/PST045222}--Asian: 5.9\%, Black: 12.6\%, Hispanic: 18.9\%, White: 59.3\%.}. Table \ref{t:ppp_racial_distribution} provides a breakdown. In the appendix, I also show that my results are robust to the original sample.

\begin{table}[h]
    \caption{PPP Racial Distribution}
    \label{t:ppp_racial_distribution}
    \centering
    \begin{tabular}{lll}
\toprule
Race & Total & Sampled \\
\midrule
Asian & 57,303 & 12,202 \\
Black & 559,667 & 26,059 \\
Hispanic & 145,913 & 39,089 \\
White & 449,137 & 122,647 \\
\bottomrule
\end{tabular}

\end{table}

\section{Results}

\subsection{LSTM}

If the optimal prediction for a given name is based on an empirically calculated distribution, why do we need machine learning at all? The primary reason is coverage. Due to either data limitations or privacy concerns (for example, Census finds that there are 3,899,864 last names that appear only once in their sample (\cite{census2010surnames})), the race of many names can only be inferred through machine learning. Furthermore, names that do not appear in the probability files algorithms such as BISG and BIFSG rely on often are correlated with nationality. Many African and Eastern European names, such as ``Jurczewsky'', ``Semuyaba'', and ``Ng'ethe'', to name a few, do not appear in the files. Users that rely solely on such algorithms may be unintentionally systemically excluding certain groups from their sample. 

Therefore, I follow \cite{sood2018}, \cite{kotova2021}, and \cite{fang2022} and train an LSTM to predict race from first and last name. LSTMs are a type of recurrent neural network and have shown state-of-the-art performance on many natural language processing tasks (\cite{lample2016}). Bidirectional LSTMs add a backward layer to a regular LSTM where the information is reversed, and have been shown to be able to better capture the context of text (\cite{graves2005}). 

Before training, I remove all numbers and punctuation except for hyphens, spaces, and apostrophes, which appear frequently in names and are unlikely to be data errors. I also remove observations where the first or last name is only one character. I use an 80:20 split for training and validation. After splitting, I undersample the dataset so that each class has an equal number of observations. In comparison, an imbalanced dataset could lead the model to optimize by simply predicting the majority class (White) most of the time. Then, I concatenate first and last names into full names and character-encode them. I assign each character in my vocabulary (all lowercase letters plus hyphens and apostrophes) a number, such that ``smith'' becomes [19, 13, 9, 20, 8]. I use 30 as the window size, meaning that shorter names are right padded with 0's and longer names are truncated at 30 characters. Character-level features work well with neural networks since they are good at extracting information from raw data \cite{zhang2015}. I pass the encoded vectors into an embedding layer with an embedding dimension of 256. The embeddings are fed into four BiLSTM layers with hidden size 512 and a dropout rate of 0.2, and a final dense layer with softmax activation. I use the Adam optimizer with .001 learning rate and a weight decay of .004, and use a batch size of 512. I consider alternative architectures in the appendix.

To the best of my knowledge, the resulting model has better performance than existing models in the literature. For example, my model achieves a F1 score of 0.642 for Black, compared to 0.552 for \cite{sood2018}, 0.513 for \cite{fang2022}, and 0.47 for \cite{kotova2021}. A full summary of the performance of \cite{sood2018} (ethnicolr) and \cite{fang2022} (rethnicity) on the aforementioned PPP test sample can be found in Tables \ref{t:eth_stats_max} and \ref{t:reth_stats_max}\footnote{\cite{kotova2021} does not have an associated open-source package to test against, so the number is taken directly from the paper rather than calculated on the PPP sample.}. 

\begin{figure}[H]
    \caption{F1 Comparison (First-Last vs. Ethnicolr vs. Rethnicity)}
    \label{f:fl_eth_reth_f1_compare}
    \centering
    \includegraphics[scale=0.8]{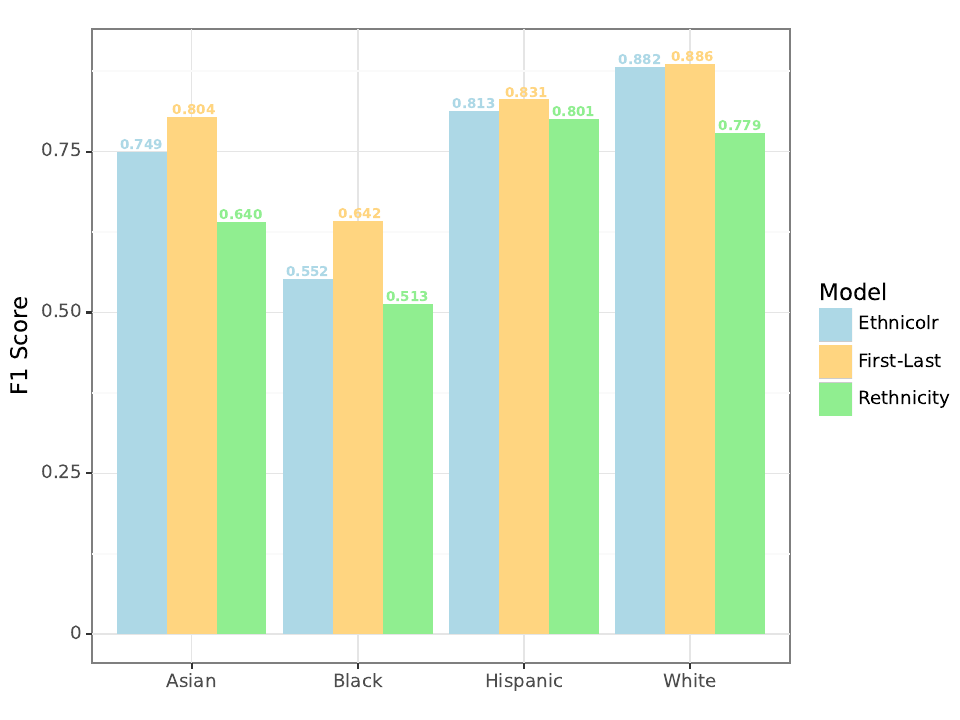}
\end{figure}

Importantly, most of the errors the model makes seem to be ``reasonable''--one could imagine a human making the same mistake with the same information. For example, the model gets ``Felicia Gray'', ``Barbara Middleton'', and ``Karen Ross'' wrong, who all self-report as Asian. Similarly, the model predicts ``Surinder Kaur'', ``Balbir Ghandi'', and ``Tu Vuong'' as Asian, but they self-report as White. Table \ref{t:fl_stats_max} summarizes its performance.

\begin{table}[H]
    \caption{First-Last Stats (Max)}
    \label{t:fl_stats_max}
    \centering
    \begin{tabular}{lllllll}
\toprule
Race & Accuracy & Precision & Recall & F1 Score & Coverage & Support \\
\midrule
Asian & 0.975 & 0.772 & 0.839 & 0.804 & 1.0 & 12,202 \\
Black & 0.898 & 0.589 & 0.705 & 0.642 & 1.0 & 26,059 \\
Hispanic & 0.937 & 0.871 & 0.795 & 0.831 & 1.0 & 39,089 \\
White & 0.862 & 0.896 & 0.876 & 0.886 & 1.0 & 122,647 \\
\bottomrule
\end{tabular}

\end{table}

Location also encodes important information. Instead of adding location features to the model, which may not be portable (if, for example, a user wants to use tract-level instead of ZCTA-level data), I use naive Bayes:

\begin{align*}
    P(r | n, g) = \frac{P(r | n) \times P(g | r)}{\sum_{r=1}^{4} P(r | n) \times P(g | r)}
\end{align*},

where \( n \) is name, \( r \) is race (Asian, Black, Hispanic, White), \( g \) is geography (in this case, ZCTA), and \( P(r | n) \) are the probabilities returned by the aforementioned name-only model. The resulting model achieves similar results for Asian and Hispanic, but makes a significant leap for Black (a 17\% increase in F1 score) and a modest gain for White (a 3\% increase in F1 score). This makes sense, as Black and White names are the most likely to be confused for each other, and is where location features can make the most difference. For instance, a person with the surname ``Li'' is likely Asian regardless of location, and indeed location may even just add noise in such cases.

\begin{table}[H]
    \caption{First-Last-ZCTA Stats (Max)}
    \label{t:flz_bayes_stats_max}
    \centering
    \begin{tabular}{lllllll}
\toprule
Race & Accuracy & Precision & Recall & F1 Score & Coverage & Support \\
\midrule
Asian & 0.974 & 0.744 & 0.864 & 0.799 & 0.994 & 12,130 \\
Black & 0.925 & 0.664 & 0.854 & 0.747 & 1.0 & 26,046 \\
Hispanic & 0.941 & 0.881 & 0.805 & 0.841 & 0.999 & 39,052 \\
White & 0.893 & 0.934 & 0.888 & 0.911 & 1.0 & 122,601 \\
\bottomrule
\end{tabular}

\end{table}

\begin{figure}[H]
    \caption{AUC Comparison (First-Last vs. First-Last-ZCTA)}
    \label{f:fl_flz_auc_compare}
    \centering
    \includegraphics[scale=.8]{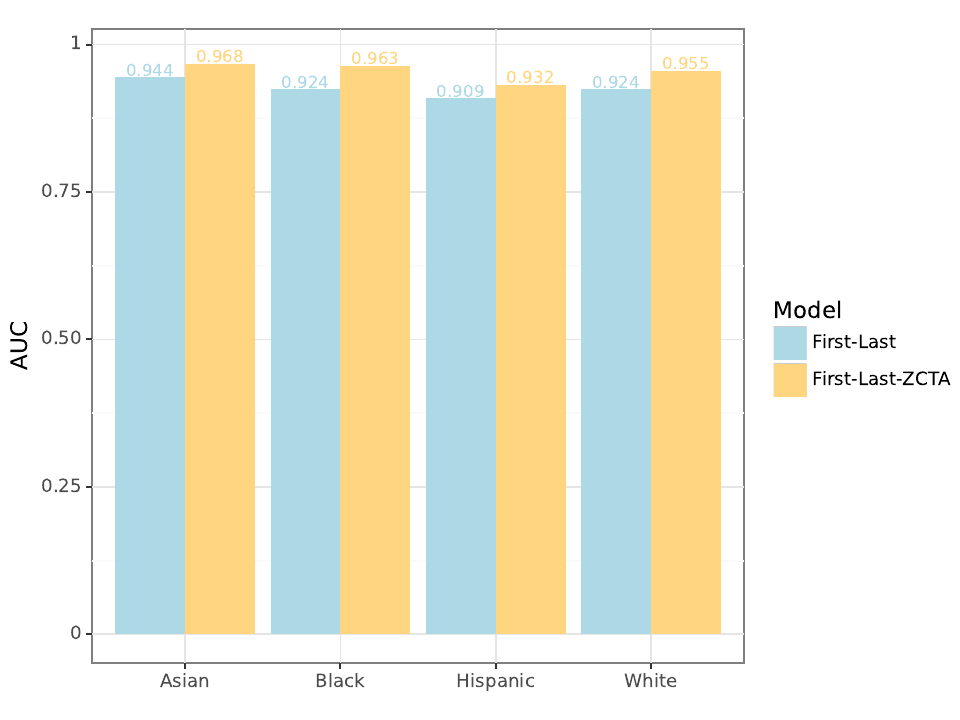}
\end{figure}

Figure \ref{f:fl_flz_auc_compare} compares the area under the ROC curve (AUC) scores when ZCTA features are added. Again, the gains are the greatest for Black and White, although the increases are significant across the board.

\subsection{Ensemble}

While BiLSTM shows very good performance (\cite{hosmer2000} suggest that a .9 AUC is ``excellent'') and perfect coverage, if an empirically calculated distribution (or something very close to it) is possible, that information should be incorporated. In their analysis of race prediction algorithms, \cite{rieke2022} argue that there is no single, ideal inference method and that ``different proxies are likely to be more or less suited to different kinds of measurement tasks.'' Although I agree with their conclusion, practically, it is difficult to know \textit{a priori} which proxy is best suited to a specific task. Furthermore, ensemble methods have been shown to be able to compensate for each individual model's weaknesses (\cite{dietterich2000}). Thus, I create a simple ensemble that combines the BiLSTM with BISG and BIFSG. I choose BISG and BIFSG because they are theoretically close to the optimal solution. As a bonus, they are also highly interpretable and computationally cheap.

\subsubsection{BISG} \label{subsubsec:bisg}

The canonical version of BISG, developed by the Rand Corporation in 2009, calculates the probability a person is of a certain race as

\begin{align*}
    P(r | s, g) = \frac{P(r | s) \times P(g | r)}{\sum_{r=1}^{6} P(r | s) \times P(g | r)}
\end{align*} 

where \( r \) is one of  American Indian or Alaska Native, Asian or Pacific Islander, Black, Hispanic, White, or Multiracial, \( s \) denotes surname, and \( g \) denotes geography. For the purposes of this paper, the geography is at the ZCTA level, although it can be defined at the census block, census tract, county, and state levels as well. \( P(g | r) \) is the percentage of people of a certain race that live in the specified geography. \( P(r | s) \) is defined as the percentage of people with a given surname that are of that race, and is calculated from the 2010 US census. The Census surname table comprises all surnames that appear more than 100 times, and yields 162,254 unique surnames covering 90\% of the US population (\cite{census2010surnames}). 

To improve coverage, I update the table with probabilities calculated from the L2 voter data. Since the L2 data is highly imbalanced, I draw a nationally representative sample, otherwise the probabilities would be artificially biased towards the majority classes. Additionally, I only consider the four major racial categories, allowing me to have substantially more observations. When building the table, I follow \cite{tzioumis2018} by deleting suffixes such as ``JR'', names that are only one character long, deleting blanks and hyphens, and only considering names that either have 30 or more observations or names that have 15-29 observations and represent one and only one race. If two names overlap between Census and L2, I prefer the Census values. Although the Census probabilities are based on older data, they come from over 230 million observations (\cite{census2010surnames}), compared to 51,294,450 for L2. Additionally, the Census is a random draw from the US population, whereas the L2 data only represents registered voters. I refer to this combined version as ``iBISG.'' Tables \ref{t:bisg_stats_max} and \ref{t:ibisg_stats_max} summarize their performances.

\begin{table}[H]
    \caption{BISG Stats (Max)}
    \label{t:bisg_stats_max}
    \centering
    \begin{tabular}{lllllll}
\toprule
Race & Accuracy & Precision & Recall & F1 Score & Coverage & Support \\
\midrule
Asian & 0.986 & 0.904 & 0.842 & 0.872 & 0.875 & 10,674 \\
Black & 0.915 & 0.689 & 0.675 & 0.682 & 0.947 & 24,673 \\
Hispanic & 0.943 & 0.913 & 0.794 & 0.849 & 0.947 & 37,008 \\
White & 0.882 & 0.882 & 0.93 & 0.905 & 0.902 & 110,663 \\
\bottomrule
\end{tabular}

\end{table}

\begin{figure}[H]
    \caption{F1 Comparison (First-Last-ZCTA vs. iBISG)}
    \label{f:flz_ibisg_f1_compare}
    \centering 

    \includegraphics[scale=0.8]{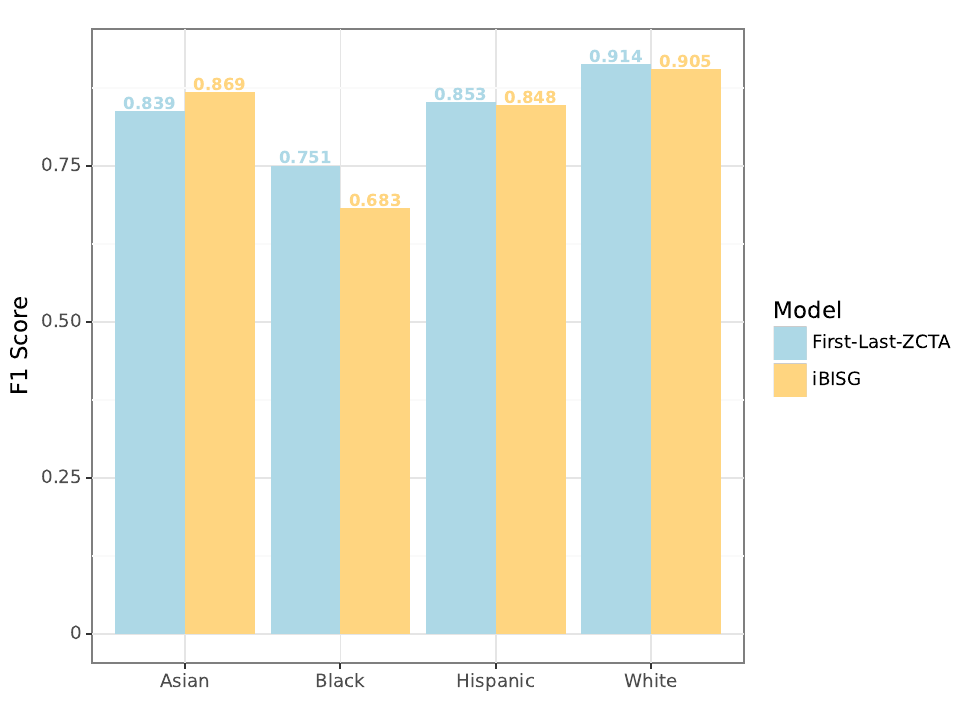}

    \begin{minipage}{\textwidth} 
        \medskip
		\footnotesize{
            \textbf{Note:} I subset to where all models are able to make a prediction.
        }
	\end{minipage}
\end{figure}

\begin{table}[H]
    \caption{iBISG Stats (Max)}
    \label{t:ibisg_stats_max}
    \centering
    \begin{tabular}{lllllll}
\toprule
Race & Accuracy & Precision & Recall & F1 Score & Coverage & Support \\
\midrule
Asian & 0.985 & 0.899 & 0.842 & 0.869 & 0.885 & 10,798 \\
Black & 0.915 & 0.69 & 0.676 & 0.683 & 0.951 & 24,777 \\
Hispanic & 0.943 & 0.91 & 0.794 & 0.848 & 0.951 & 37,176 \\
White & 0.882 & 0.882 & 0.929 & 0.905 & 0.908 & 111,314 \\
\bottomrule
\end{tabular}

\end{table}

iBISG shows modest improvements in coverage while showing virtually unchanged performance. This is not surprising--the Census Bureau should have the highest quality data, and already covers the majority of surnames in the US. But what does iBISG add to the ensemble? Figure \ref{f:flz_ibisg_f1_compare} compares the F1 scores of First-Last-ZCTA and iBISG on the sample where iBISG is able to make predictions. Note that comparing the F1 scores directly is misleading because names where iBISG is unable to make predictions are, by definition, rarer, and may be more difficult for the model.

Figure \ref{f:flz_ibisg_f1_compare} shows that iBISG tends to make more accurate predictions for Asians. Asians often have quite distinctive last names that have little overlap with other classes, so a surname-only approach's advantages make sense here. It is possible that the model sometimes mispredicts when the first name is another class (for instance, many Asians who immigrate to the US will change their first names to something more ``Western'' but keep their last names the same).

\subsubsection{BIFSG} \label{subsec:bifsg}

\cite{voicu2018} offers an extension of BISG in BIFSG, a similar algorithm that incorporates first name data. BIFSG calculates the probability a person is of a certain race as

\begin{align*}
    P(r | s, f, g) = \frac{P(r | s) \times P(f | r) \times P(g | r)}{\sum_{r=1}^{6} P(r | s) \times P(f | r) \times P(g | r)}
\end{align*}

All variables are defined in the same way as in BISG. Theoretically, first name data should improve performance, especially for Black and White, as they tend to have distinct first names but similar surnames (\cite{bertrand2004}). In practice, the gains often either do not materialize or are offset by regressions in other metrics. For instance, \cite{pace2022} finds that BIFSG improves Black F1 score by only 0.04 and that ``the non-random nature of first name exclusions may result in the counter-intuitive decrease in BIFSG/BISG aggregate predictive accuracy for one or more groups.'' I find similar results. Table \ref{t:bifsg_stats_max} shows the performance of BIFSG on the PPP test sample.

\begin{table}[H]
    \caption{BIFSG Stats (Max)}
    \label{t:bifsg_stats_max}
    \centering
    \begin{tabular}{lllllll}
\toprule
Race & Accuracy & Precision & Recall & F1 Score & Coverage & Support \\
\midrule
Asian & 0.989 & 0.942 & 0.823 & 0.879 & 0.589 & 7,187 \\
Black & 0.941 & 0.7 & 0.616 & 0.656 & 0.536 & 13,957 \\
Hispanic & 0.949 & 0.928 & 0.797 & 0.858 & 0.764 & 29,881 \\
White & 0.904 & 0.902 & 0.961 & 0.931 & 0.843 & 103,437 \\
\bottomrule
\end{tabular}

\end{table}

Compared to BISG, I find that, for Black, BIFSG has 43.4\% worse coverage and a 3.8\% lower F1 score. However, BIFSG's performance issues are due primarily to data issues. The existing firm name list comes from \cite{tzioumis2018}, who uses Home Mortgage Disclosure Act (HMDA) data. The list consists of 2,663,364 observations representing 91,526 unique first names. I update the list with the same L2 data as described in \ref{subsubsec:bisg}, yielding 51,294,450 observations representing 1,187,576 unique first names. Since the HMDA data has fewer observations, I prefer the L2 values if they exist. I refer to this combined version as ``iBIFSG.'' Table \ref{t:bifsg_stats_max} summarizes its performance.

\begin{table}[H]
    \caption{iBIFSG Stats (Max)}
    \label{t:ibifsg_stats_max}
    \centering
    \begin{tabular}{lllllll}
\toprule
Race & Accuracy & Precision & Recall & F1 Score & Coverage & Support \\
\midrule
Asian & 0.988 & 0.905 & 0.873 & 0.889 & 0.764 & 9,325 \\
Black & 0.936 & 0.713 & 0.766 & 0.738 & 0.786 & 20,487 \\
Hispanic & 0.947 & 0.911 & 0.813 & 0.859 & 0.881 & 34,433 \\
White & 0.908 & 0.917 & 0.939 & 0.928 & 0.887 & 108,832 \\
\bottomrule
\end{tabular}

\end{table}

In this case, iBIFSG shows significant improvements in coverage across all classes, with the greatest gains coming from minority classes. In particular, for Black, iBIFSG achieves a 12.5\% increase in F1 Score and a 46.6\% increase in coverage compared to BIFSG. The gains are mainly due to increased recall. That is to say, iBIFSG correctly identifies 76.6\% of all Black borrowers in the sample, compared to just 61.6\% for BIFSG.

Furthermore, on a subset where all of iBIFSG, iBISG, and First-Last-ZCTA are able to make a prediction, iBIFSG has the highest F1 scores across the board. However, when considering the full sample, compared to First-Last-ZCTA, it still has 13.4\% worse coverage overall and up to 23.1\% worse coverage for specific classes.

\begin{figure}[H]
    \caption{F1 Comparison (First-Last-ZCTA vs. iBISG vs. iBIFSG)}
    \label{f:flz_ibisg_ibifsg_f1_compare}
    \centering 

    \includegraphics[scale=.7]{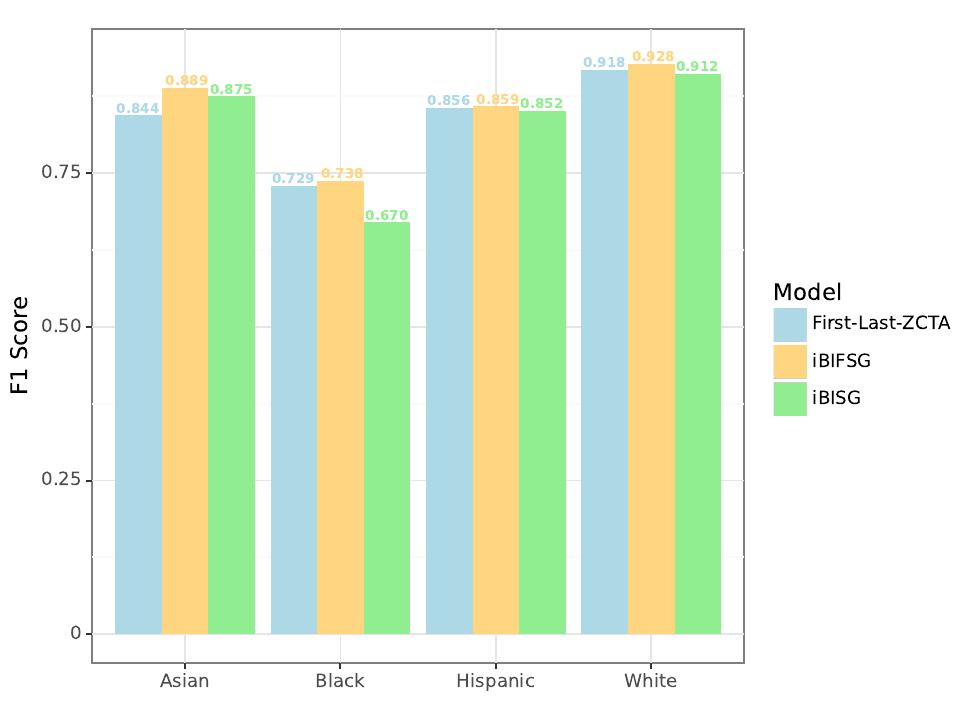}

    \begin{minipage}{\textwidth} 
        \medskip
		\footnotesize{
            \textbf{Note:} I subset to where all models are able to make a prediction.
        }
	\end{minipage}
\end{figure}

\subsubsection{Final Ensemble}

In this section, I present a simple ensemble model that maximizes coverage while maintaining high performance. I take the weighted average of the predictions made by iBIFSG, iBISG, and First-Last-ZCTA, assigning equal weight to each model that is able to make a prediction. Table \ref{t:ensemble_stats_max} reports the performance.

\begin{table}[H]
    \caption{Ensemble Stats (Max)}
    \label{t:ensemble_stats_max}
    \centering
    \begin{tabular}{lllllll}
\toprule
Race & Accuracy & Precision & Recall & F1 Score & Coverage & Support \\
\midrule
Asian & 0.979 & 0.803 & 0.862 & 0.831 & 1.0 & 12,202 \\
Black & 0.931 & 0.706 & 0.811 & 0.755 & 1.0 & 26,059 \\
Hispanic & 0.943 & 0.895 & 0.802 & 0.846 & 1.0 & 39,089 \\
White & 0.9 & 0.921 & 0.915 & 0.918 & 1.0 & 122,647 \\
\bottomrule
\end{tabular}

\end{table}

\begin{figure}[H]
    \caption{F1 Comparison}
    \label{f:all_f1_compare}
    \centering
    \includegraphics[scale=.8]{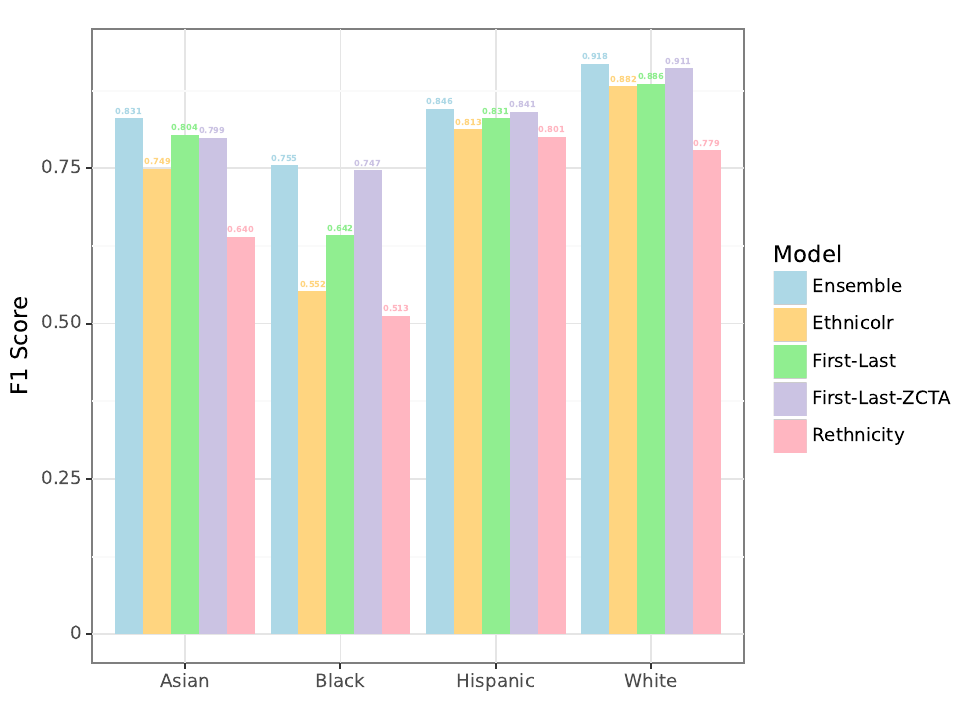}
\end{figure}

The ensemble reports higher F1 scores across the board than First-Last-ZCTA and maintains perfect coverage. In the future, more sophisticated weighting schemes may improve performance. For example, it is possible that certain models perform well in certain geographies, and geography-specific weights could be constructed. However, since the PPP sample does not cover all ZCTAs, and does not have a statistically significant number of observations for every ZCTA, I do not attempt this exercise. However, even with the most basic ensembling strategy, the final model outperforms all publicly available models I was able to test. Figure \ref{f:all_f1_compare} provides a comparison of the F1 scores. The gains are especially apparent for Asian and Black. Compared to the next-best performing model in the literature, ethnicolr, the ensemble achieves a 10.9\% increase in F1 score for Asians, a 36.8\% increase for Blacks, and a 4.1\% increase for Hispanics and Whites. This pattern holds true for a random (instead of nationally representative) draw of 200,000 observations from the full PPP sample, where Black is the majority class, suggesting that this approach is robust to alternative sample constructions, and does not tend to unfairly overpredict one class versus another. In general, First-Last and First-Last-ZCTA perform very well, although the benefit of the ensemble is clear for Asians. Additionally, iBISG and iBIFSG are easy to implement and run very quickly, so it is virtually zero-cost to combine them with any machine learning model.

The same holds true when comparing to BISG on the sample where BISG is able to make a prediction, especially for Black (Figure \ref{f:bisg_all_f1_compare}). While the primary benefit of a machine learning model is to be able to make predictions for any name, it is important that performance does not regress on the subset where the current standard algorithm is able to make predictions.

\section{Discussion}

Can we trust race prediction? Even with perfect information, using only name and location inherently limits the accuracy of available models. For instance, a Black person with an ambiguous name (``Mary Williams'', ``Ashley Jackson'') living in New York City will likely always be mislabeled by such models as White. Similarly, many Filipinos with Hispanic-sounding names (such as ``Maria Cruz Santos'') will be labeled as Hispanic. However, it is clear that models have the ability to achieve very high performance. iBIFSG and the resulting ensemble already show significant improvements over existing solutions that have been used in such critical contexts as fair lending analysis. On a nationally representative sample of PPP data, the ensemble achieves perfect coverage and up to 36.8\% better F1 scores than ethnicolr, the current leading machine learning model. In the future, improvements such as more data or advances in architectures could further push performance. For instance, using XGBoost instead of a simple average to ensemble the models may be more robust. Additionally, voter data is inherently not representative, as not everybody is registered to vote, especially Blacks and Hispanics (\cite{sood2018}).

\clearpage

\singlespacing
\setlength\bibsep{0pt}
\bibliographystyle{abbrvnat}
\bibliography{race_ml_bib}

\clearpage

\setcounter{table}{0} \renewcommand{\thetable}{A.\arabic{table}}											
\setcounter{figure}{0} \renewcommand{\thefigure}{A.\arabic{figure}} \setcounter{page}{1}							
\setcounter{section}{0} \renewcommand\thesection{\Alph{section}}											
\setcounter{page}{0} \renewcommand\thepage{\arabic{page}}												
\fancypagestyle{alim}{\fancyhf{}\renewcommand{\headrulewidth}{0pt}\fancyfoot[C]{Appendix \thepage}}		
\pagestyle{alim}


\vspace*{\fill}
\thispagestyle{empty}
\begin{center}
\textcolor{white}{}\\
\textbf{\textcolor{black}{\LARGE{} Appendix}}{\Large\par}
\par\end{center}

\begin{center}
\par\end{center}
\vspace*{\fill}

\clearpage
\onehalfspacing
\section*{Appendix Figures} \label{sec:appendix_figures}
\addcontentsline{toc}{section}{Tables}

\begin{figure}[H]
    \caption{F1 Comparison (BISG sample)}
    \label{f:bisg_all_f1_compare}
    \centering
    \includegraphics[scale=0.8]{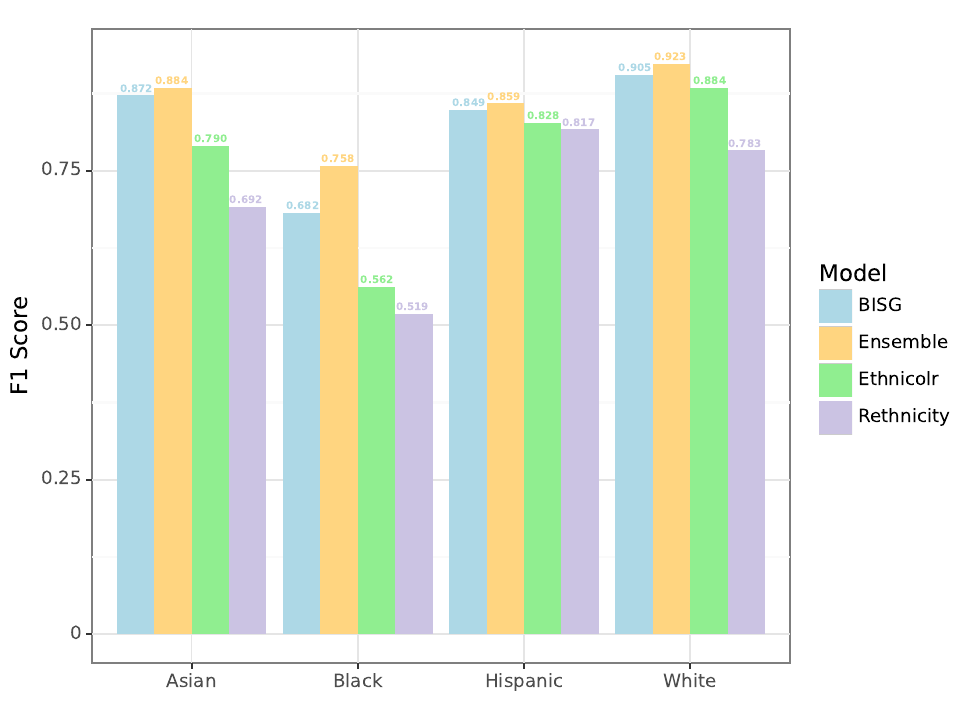}

    \begin{minipage}{\textwidth} 
        \medskip
		\footnotesize{
            \textbf{Note:} I subset to where all models are able to make a prediction.
        }
	\end{minipage}
\end{figure}

\begin{figure}[H]
    \caption{F1 Comparison (PPP Random)}
    \label{f:ppp_rand_all_f1_compare}
    \centering
    \includegraphics[scale=0.8]{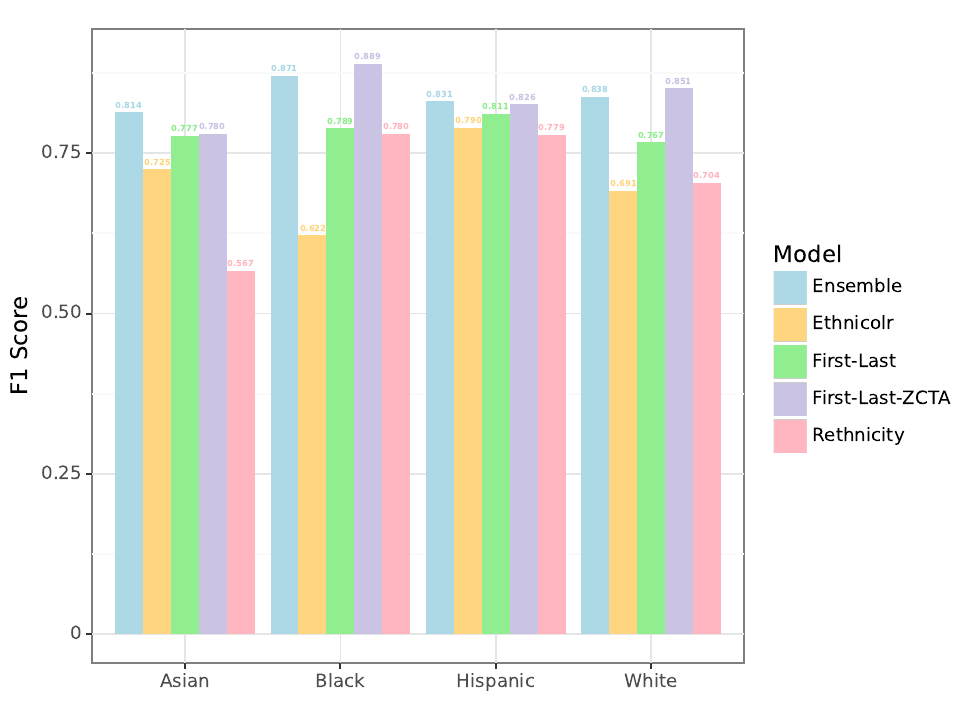}

    \begin{minipage}{\textwidth} 
        \medskip
		\footnotesize{
            \textbf{Note:} I use a random (instead of nationally representative) draw of 200,000 observations from the full PPP sample.
        }
	\end{minipage}
\end{figure}

\clearpage
\onehalfspacing
\section*{Appendix Tables} \label{sec:appendix_tables}
\addcontentsline{toc}{section}{Tables}

\begin{table}[H]
    \caption{Rethnicity Stats (Max)}
    \label{t:reth_stats_max}
    \centering
    \begin{tabular}{lllllll}
\toprule
Race & Accuracy & Precision & Recall & F1 Score & Coverage & Support \\
\midrule
Asian & 0.939 & 0.502 & 0.882 & 0.64 & 1.0 & 12,202 \\
Black & 0.811 & 0.386 & 0.763 & 0.513 & 1.0 & 26,059 \\
Hispanic & 0.926 & 0.845 & 0.761 & 0.801 & 1.0 & 39,089 \\
White & 0.763 & 0.909 & 0.682 & 0.779 & 1.0 & 122,647 \\
\bottomrule
\end{tabular}

\end{table}

\begin{table}[H]
    \caption{Ethnicolr Stats (Max)}
    \label{t:eth_stats_max}
    \centering
    \begin{tabular}{lllllll}
\toprule
Race & Accuracy & Precision & Recall & F1 Score & Coverage & Support \\
\midrule
Asian & 0.974 & 0.897 & 0.643 & 0.749 & 1.0 & 12,202 \\
Black & 0.901 & 0.673 & 0.469 & 0.552 & 1.0 & 26,059 \\
Hispanic & 0.932 & 0.882 & 0.754 & 0.813 & 1.0 & 39,089 \\
White & 0.845 & 0.828 & 0.943 & 0.882 & 1.0 & 122,647 \\
\bottomrule
\end{tabular}

\end{table}

\clearpage


\clearpage

\end{document}